\pdfoutput=1

\documentclass[11pt]{article}

\usepackage[preprint]{acl}

\usepackage{times}
\usepackage{latexsym}

\usepackage[T1]{fontenc}

\usepackage[utf8]{inputenc}

\usepackage{microtype}

\usepackage{inconsolata}

\usepackage{graphicx}
\usepackage{multirow}
\usepackage{booktabs}
\usepackage{tikz}
\usepackage{xcolor}
\usepackage{amsmath}
\usepackage{amssymb}
\usepackage{graphicx}
\usepackage{makecell}
\usepackage{xspace}
\usepackage{arydshln}
\usepackage{tcolorbox}
\usepackage{tabularx}

\newcommand{\garage}{GaRAGe}

\title{GaRAGe: A Benchmark with Grounding Annotations \\ for RAG Evaluation}

\author{Ionut-Teodor Sorodoc
\\
  Amazon AGI \\
  \texttt{csorionu@amazon.com} \\\And
  Leonardo F. R. Ribeiro \\
  Amazon AGI \\
  \texttt{leonribe@amazon.com} \\\AND
  Rexhina Blloshmi \\
  Amazon AGI \\ 
  \texttt{blloshmi@amazon.com} \\\And
  Christopher Davis \\
  Amazon AGI \\
  \texttt{davisjnh@amazon.co.uk} \\\And
  Adrià de Gispert \\
  Amazon AGI \\
  \texttt{agispert@amazon.com} \\}

\begin{document}
\maketitle
\begin{abstract}
We present \garage{}, a large RAG benchmark with human-curated long-form answers and annotations of each grounding passage, allowing a fine-grained evaluation of whether LLMs can identify relevant grounding when generating RAG answers. Our benchmark contains 2366 questions of diverse complexity, dynamism, and topics, and includes over 35K annotated passages retrieved from both private document sets and the Web, to reflect real-world RAG use cases. This makes it an ideal test bed to evaluate an LLM's ability to identify only the relevant information necessary to compose a response, or provide a deflective response when there is insufficient information. 
Evaluations of multiple state-of-the-art LLMs on \garage{} show that the models tend to over-summarise rather than (a) ground their answers strictly on the annotated relevant passages (reaching at most a {\it Relevance-Aware} Factuality Score of 60\%), or (b) deflect when no relevant grounding is available (reaching at most 31\% true positive rate in deflections). The F$_{1}$ in attribution to relevant sources is at most 58.9\%, and we show that performance is particularly reduced when answering time-sensitive questions and when having to draw knowledge from sparser private grounding sources.\footnote{The dataset can be found at \url{https://github.com/amazon-science/GaRAGe}}

\end{abstract}

\begin{table*}[h]

    \centering
    \resizebox{\textwidth}{!}{
    \renewcommand{\arraystretch}{1}
    \begin{tabular}{lccccccccccccc}
        \toprule
        & \multicolumn{2}{c}{\textbf{Human Intervention}} & \multicolumn{3}{c}{\textbf{Question}} & \multicolumn{3}{c}{\textbf{Answer}} & \multicolumn{3}{c}{\textbf{Grounding}} \\
        \cmidrule(r){2-3} \cmidrule(r){4-6} \cmidrule(r){7-9} \cmidrule(r){10-12}
        & \makecell{Human \\ Validation} & \makecell{Human \\ Annotation} & \makecell{Temporal \\ Dynamism} & \makecell{Complexity \\ Variation} & \makecell{Detailed \\Annotation} & Comprehensive & \makecell{Contain \\ Citations} & Deflection & \makecell{Public \\ and Private} & \makecell{Annotated \\ Grounding} & \makecell{Contain \\Metadata} \\
        \midrule
        MultiHop RAG \citep{tang2024multihoprag} & \textcolor{red}{$\times$} & \textcolor{red}{$\times$} & \checkmark & $\cdot$ & \checkmark & \textcolor{red}{$\times$} & \textcolor{red}{$\times$} & \checkmark & \textcolor{red}{$\times$} & $\cdot$ & \checkmark \\
        CRAG \citep{yang2024crag} & \checkmark & $\cdot$ & \checkmark & \checkmark & \checkmark & \textcolor{red}{$\times$} & \textcolor{red}{$\times$} & \checkmark & $\cdot$ & \textcolor{red}{$\times$} & \checkmark \\
        ConcurrentQA \citep{arora2023reasoning} & \checkmark & \checkmark & \textcolor{red}{$\times$} & \checkmark & \textcolor{red}{$\times$} & \textcolor{red}{$\times$} & \textcolor{red}{$\times$} & \textcolor{red}{$\times$} & \checkmark & $\cdot$ & \textcolor{red}{$\times$} \\
        Summary of a Haystack \citep{laban-etal-2024-summary} & $\cdot$ & $\cdot$ & \textcolor{red}{$\times$} & $\cdot$ & \textcolor{red}{$\times$} & \checkmark & \checkmark & \textcolor{red}{$\times$} & \textcolor{red}{$\times$} & \textcolor{red}{$\times$} & \checkmark \\
        RAG-QA Arena \citep{han2024rag} & \checkmark & \checkmark & $\cdot$ & $\cdot$ & $\cdot$ & \checkmark & \checkmark & \textcolor{red}{$\times$} & \textcolor{red}{$\times$} & \textcolor{red}{$\times$} & \checkmark \\
        Facts Grounding \cite{jacovi2024facts} & \checkmark & \checkmark & \textcolor{red}{$\times$} & $\cdot$ & \checkmark & \checkmark & \textcolor{red}{$\times$} & \textcolor{red}{$\times$} & \textcolor{red}{$\times$} & \textcolor{red}{$\times$} & \textcolor{red}{$\times$} \\
        \textbf{GaRAGe} (ours) & \checkmark & \checkmark & \checkmark & \checkmark & \checkmark & \checkmark & \checkmark & \checkmark & \checkmark & \checkmark & \checkmark \\
        \bottomrule
    \end{tabular}}
    \caption{Features of the GaRAGe benchmark relative to existing RAG benchmarks.}
    \label{tab:comparison}
\end{table*}

\section{Introduction}

\begin{figure}[t]
    \centering
    \includegraphics[width=\columnwidth]{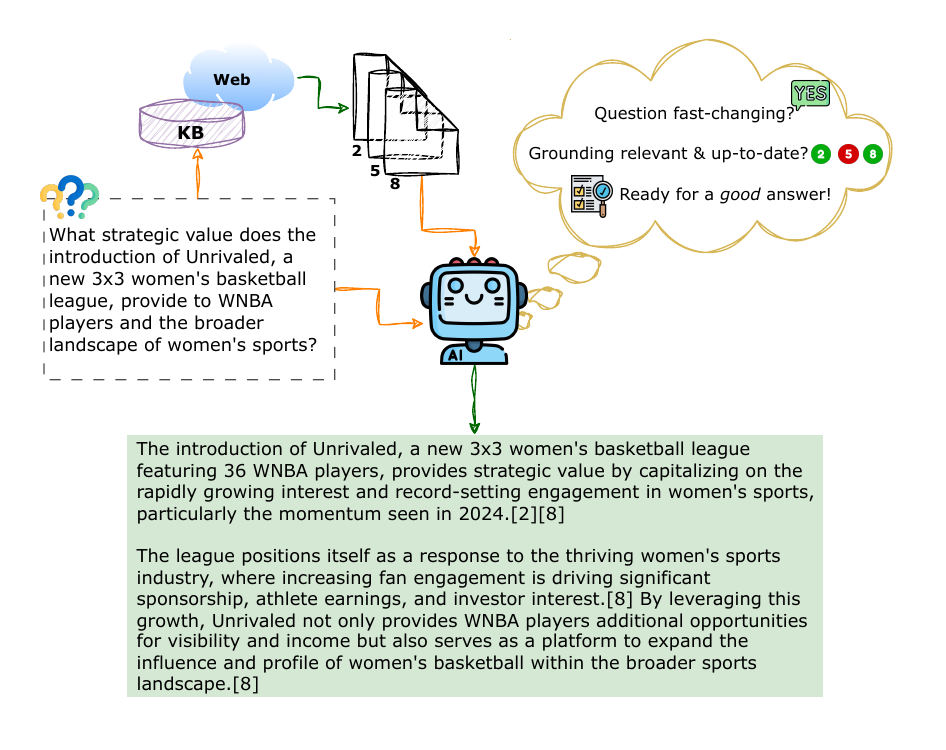}
    \caption{Example of GaRAGe Q\&A datapoint with a complex question and human curated long answers.}
    \label{fig:example}
\end{figure}
Large Language Models (LLMs) have shown consistent improvements across many tasks requiring natural language understanding, coding, mathematical or logical reasoning and have been widely incorporated in user applications, especially in the form of retrieval augmented generation (RAG) systems~\cite{NEURIPS2020_6b493230}. RAG is crucial in real-world applications, where users (a) have their own private documents to search through, and (b) require responses to be grounded on relevant, timely and attributable sources. 
For RAG applications to be successful at satisfying these needs, LLMs must simultaneously exhibit strong performance at two core competencies: query generation and answer generation. 
For query generation, LLMs must be able to determine when to invoke the available external sources and generate the right query, or queries, that will likely return good search results (grounding information), conditioned on the available (conversational, situational, visual) context.
In answer generation, LLMs must be able to provide a fluent, helpful and possibly long-form response to the user question that is grounded on the available information. Crucially, this task is not fulfilled with a simple summarisation step, because in reality the available grounding, obtained from either the user's private documents or the Web, will contain a mixture of relevant and irrelevant passages~\cite{oh-thorne-2023-detrimental}. Therefore, the LLM needs to analyse the relevancy of each grounding snippet prior to generating the answer, which will typically include citation markers to the selected source documents for transparency and accountability. An example that reflects this process is presented in Figure \ref{fig:example}.

The process of generating an accurate final answer remains crucial, regardless of the specific approach to query generation. Whether utilizing single or multiple queries, implementing one or several passes, incorporating reasoning chains, or drawing from various information sources, the fundamental challenge persists: the LLM must carefully distill relevant information from noise passages to provide precise, factual responses that address user needs while avoiding hallucinations.

Although there exist a large number of RAG benchmarks in the literature, reliably evaluating each of these competencies remains a challenge, because query generation affects the available grounding and this process lacks human annotations. Previous works resort to either: evaluating the final response directly \citep{wei2024measuring}, which conflates query generation and answer generation; or using fixed grounding from one or more documents whose relevancy is either guaranteed \citep{tang2024multihoprag, jacovi2024facts}, unknown \citep{yang2024crag} or synthetically mixed \citep{chen2024benchmarking, vodrahalli2024michelangelo}, missing the nuances of real-life retrieval from user databases and the Web. In addition to this, they come with at least one of the following disadvantages, (1) unsupervised LLM-based dataset creation \cite{tang2024multihoprag, niu-etal-2024-ragtruth, laban-etal-2024-summary} raising concerns regarding question naturalness and answer reliability, (2) short or multi-choice exam-type answers \cite{tang2024multihoprag, yang2024cragcomprehensiverag, guinet2024automated} for simpler evaluation at the cost of departing from real user expectations, (3) automatic evaluations of faithfulness to the {\it un-annotated} grounding which disregard whether the grounding is correct (e.g. \citeauthor{jacovi2024facts}, \citeyear{jacovi2024facts}), and (4) do not consider mixed grounding obtained from both user-specific document sets and the Web.

To overcome these limitations, we introduce \garage{}, a benchmark with Grounding Annotations for RAG evaluation, comprising 2.4K questions with over 35K manually-annotated grounding passages retrieved from both private documents and the Web. Fully annotated and validated by professional annotators, \garage{} enables a fine-grained evaluation of LLMs answer generation capabilities, for different question types (different dynamism, complexity, etc.) and with controlled degrees of grounding noise. Apart from grounding relevancy annotations for each snippet, the benchmark categorises questions along multiple dimensions: time-sensitivity, topic popularity, complexity or domain, and provides manually-crafted long-form, comprehensive responses that contain attribution to the relevant grounding snippets and reflect the expectations of RAG users nowadays.
Table \ref{tab:comparison} shows a comparison of \garage{} to other RAG benchmarks. 

 \begin{figure*}[t!]
    \centering
    \includegraphics[width=.95\textwidth]{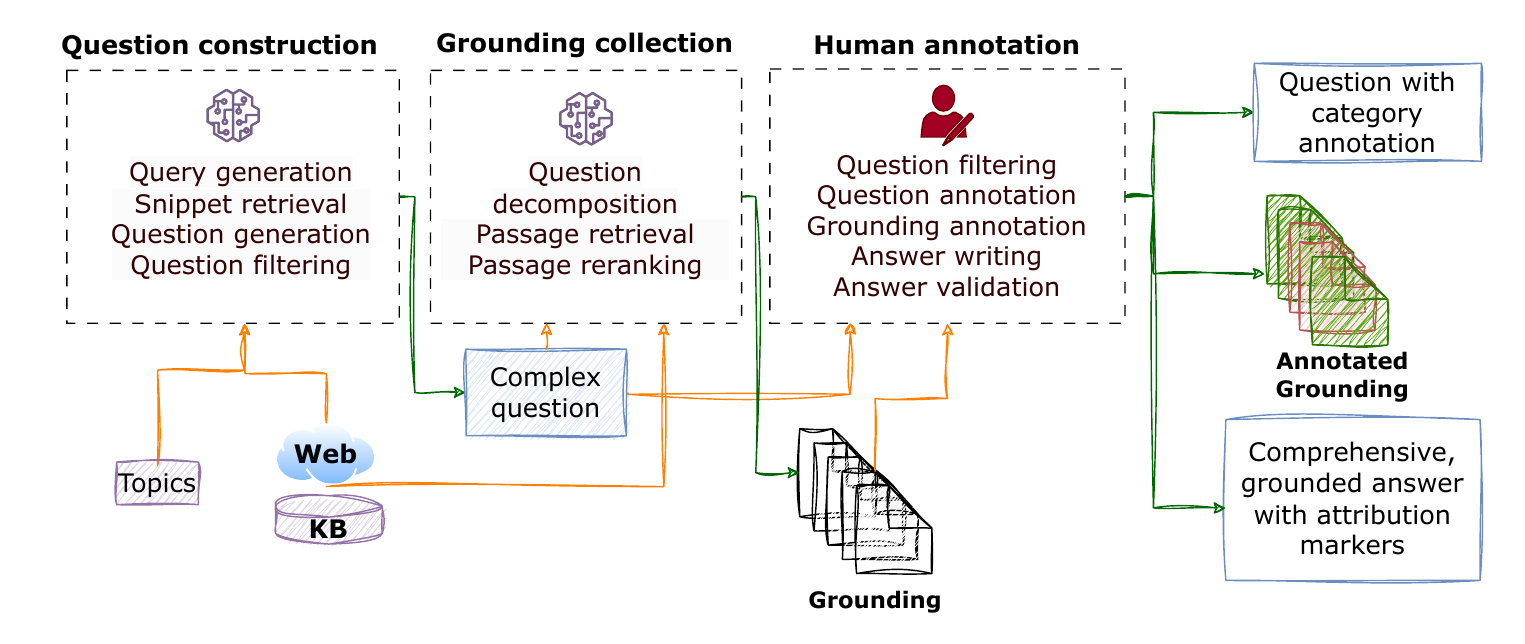}
    \caption{Pipeline for constructing \garage{} contains mainly 3 steps: (i) complex question generation through a multi-stage framework, (ii) collection of diverse grounding snippets related to the query, and (iii) human curated long-form answer creation.
    }
    \label{fig:garage_pipeline}
    
\end{figure*}
Thanks to the grounding annotations, \garage{} allows defining the Relevance-Aware Factuality Score (\textbf{RAF}), which measures whether LLM responses are grounded on \textit{strictly} relevant passages, as opposed to summarising grounding of unknown value -- extending factuality metrics from \citet{jacovi2024facts}. Furthermore, a subset of the dataset addresses the deflection capabilities of LLMs, where the provided grounding is annotated as insufficient to lead to a good answer, and the LLM is expected to avoid hallucination. Additionally, we report attribution metrics, leveraging our fine-grained annotations.
Using our benchmark and proposed metrics, we evaluate a diverse selection of state-of-the-art proprietary and open-weights LLMs and find that all models struggle at correctly detecting relevant grounding and overly summarise, reaching only 60\% in RAF, and in deflecting when the grounding is irrelevant, reaching at most 31\% true positive rate. These problems become more evident when retrieving documents from sparser private knowledge bases (KB), or when having to process time-sensitive questions which require both relevancy and temporal reasoning.
We release the dataset and the prompts used for evaluation, in order to encourage the community to evaluate and improve on the dimensions featured in the benchmark.

\section{Dataset construction}
Our methodology for creating \garage{} focuses on generating questions that challenge models with high complexity and reflect real-world RAG scenarios. The dataset features diverse grounding sources and comprehensive human-curated long form answers. 
Figure~\ref{fig:garage_pipeline} illustrates the process, which comprises the following main elements: 

\paragraph{Dynamic and Complex Questions.} A diverse set of questions with varied level of temporal dynamism, complexity, and ranging both trending and tail topics. 
For temporal dynamism we construct fast-changing, slow-changing, and static type of questions. 
These different categories require distinct interactions with the grounding information and a variable leniency towards parametric knowledge. 
Additionally, a question becomes challenging for RAG when a simple extract from a retrieved document is not sufficient to answer it~\cite{gabburo-etal-2024-measuring}. 
This complexity can arise from various factors, such as when the answer demands multiple logical steps, or requires combining information from different retrieved snippets. 
Following literature~\cite{pradeep2024ragnar}, we generate questions across dimensions such as comparisons, multi-hop and post-processing (see \S\ref{sec:question_construction}).

\paragraph{Diverse Grounding Passages.} A significant amount of passages associated to each question. 
The grounding needs to reflect the topic of questions (e.g. with passages from SEC filings if the questions talks about a specific entity from this database), but also containing enough information that can be used to compose a comprehensive answers (see \S\ref{sec:grounding_collection}). 

\paragraph{Human Validation and Annotation.} Each component is annotated and validated by professional annotators with the final goal of creating a comprehensive, grounded answer that addresses the question and contains attribution markers to signal the used information sources (see \S\ref{sec:human_annotation}).

\subsection{Question Construction}
\label{sec:question_construction}
To generate complex questions, we employ an LLM using multi-pass inference to develop the following four-step pipeline: (1) generate information-seeking Web search plan, (2) search for Web resources, (3) generate questions through information blending, and finally (4) filter and refine. 

\paragraph{Step 1: Generate Search Plan.} Inspired by previous work~\cite{luo2025kbqao1agenticknowledgebase}, we prompt the LLM to generate a strategic plan for Web exploration, focusing on identifying relevant information sources. 
This planning phase involves decomposing the overall information-seeking task into specific, actionable goals, and then generating targeted search queries for each goal to maximize the relevance and diversity of retrieved information. 
We use seed topics to bootstrap the generation. 
We add the prompts used for this step in Appendix~\ref{app:pipeline_questions_prompt1}.

\paragraph{Step 2: Search the Web.} Using the queries generated in previous step, we perform a Web search for each of them. 
This retrieval process systematically collects diverse information snippets, creating a comprehensive knowledge base that serves as the foundation for question generation.

\paragraph{Step 3: Generate Questions.} 
By incorporating exemplar questions of different types~\cite{yang2024crag} as in-context learning prompts and the collected information from the previous step, the system learns to seamlessly blend information from multiple sources, while maintaining coherence and relevance, to generate diverse questions.
This step focuses on creating questions that require synthesizing information across different snippets, promoting more challenging and realistic information retrieval scenarios. 
The prompt for generating the questions is shown in Appendix~\ref{app:pipeline_questions_prompt2}.

\paragraph{Step 4: Filtering.} The fourth stage implements a rigorous filtering framework to ensure the quality and utility of the generated questions. This includes a classifier to identify well-formed queries\footnote{https://huggingface.co/Ashishkr/query\_wellformedness\_score}, filtering questions without named entities\footnote{We use \textsc{Spacy} library.}, and deduplication employing SentenceTransformer\footnote{\texttt{all-MiniLM-L6-v2}}.

 \begin{figure*}[th!]
    \centering
    \includegraphics[width=1\textwidth]{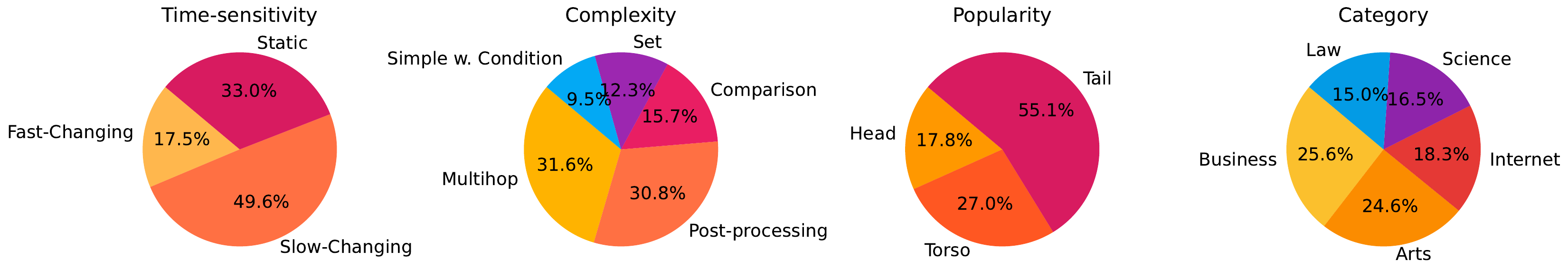}
    \caption{\textbf{Question statistics.} We report the percentage of question types for time-sensitivity, complexity, popularity and category dimensions. In total our dataset contains 2366 examples.
    }
    \label{fig:question_ann_statistics}
    
\end{figure*}

 \begin{figure}
    \centering
    \includegraphics[width=\columnwidth]{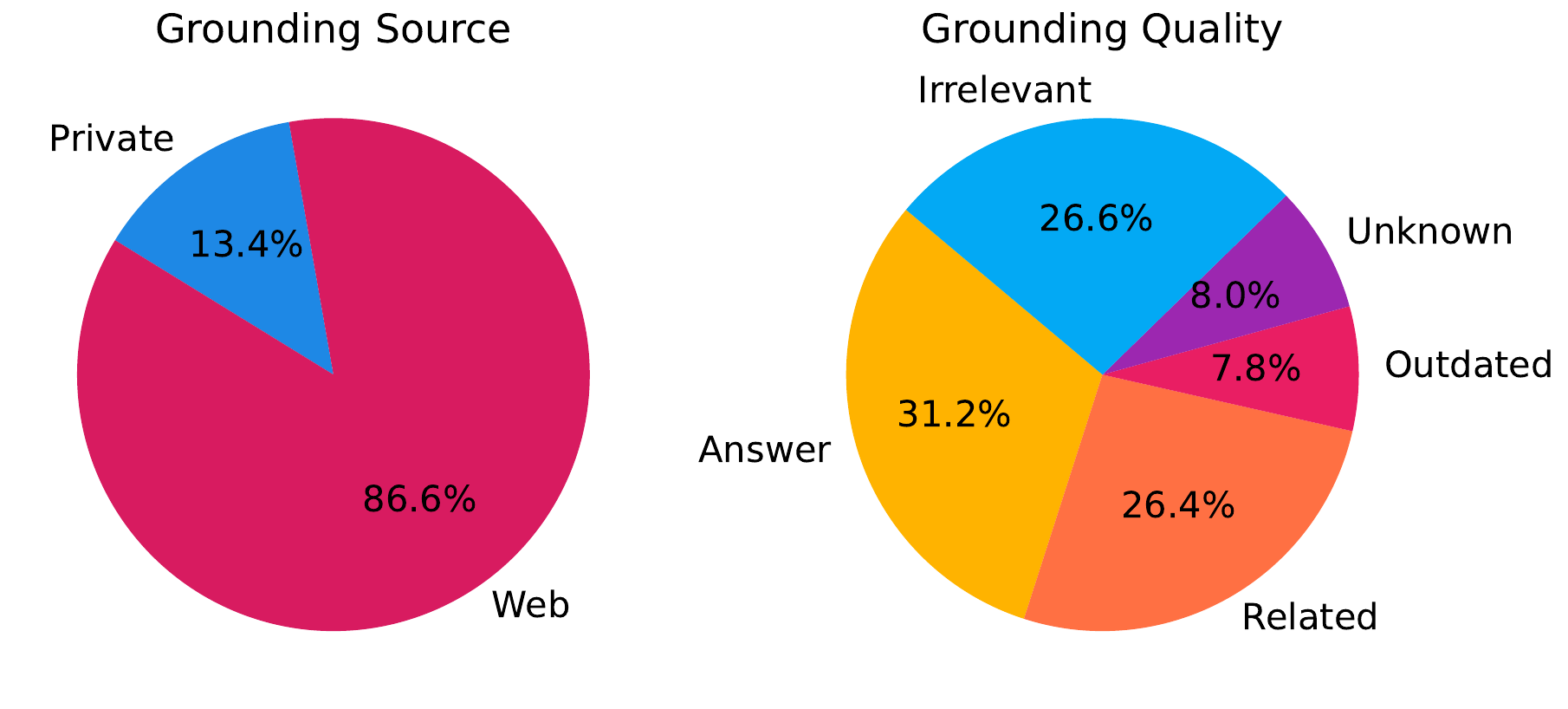}
    \caption{\textbf{Grounding statistics.} We report the percentage of grounding sources and grounding quality.
    }
    \label{fig:grounding_ans_statistics}
\end{figure}

\subsection{Grounding Collection}
\label{sec:grounding_collection}

For each of the generated questions, we collect grounding information from multiple sources. 
In particular, we utilize standalone Web search engine\footnote{To maintain anonymity, we do not disclose the name of the proprietary search engine.} or a hybrid approach that combines Web results with information retrieved from private knowledge bases. 
The private KBs that we include in this work contain Enron employee emails~\citep{klimt2004introducing}, Arxiv abstracts, AWS DevOps troubleshooting guides, and SEC filings \citep{guinet2024automated}.
The retrieval pipeline operates in two main steps. 
First, query decomposition breaks down the complex query into focused sub-queries, enabling focused retrieval of relevant information. This decomposition is performed using an LLM with a crafted prompt that encourages detailed explanations before generating the decomposition.
To maintain quality and relevance, we employ a Semantic Textual Similarity (STS) classifier based on SentenceTransformer~\cite{reimers-gurevych-2019-sentence} to filter out sub-queries that deviate from the original question's intent.

The second step involves document retrieval and reranking. 
Using the decomposed queries, we retrieve initial document sets from the selected sources. 
These documents are then reranked using a cross-encoder~\cite{chen2024bge} that computes relevance scores between the original question and each retrieved snippet. 
The final grounding consists of the top $K$ documents ranked by their cross-encoder scores, ensuring that the most relevant information is prioritized in the benchmark evaluation. Finally, we include a subset without reranking where the retrieved snippets are selected at random to include hard examples with an increased level of noise.

\subsection{Human Annotation}
\label{sec:human_annotation}
We conduct human annotation for each sample in the dataset to ensure quality. 
Professional annotators\footnote{The annotation vendor is Centrific, who provided a quote per sample of 9.65\$.} annotated 2,366 questions across four dimensions, categorised each grounding passage according to its question-relevancy, and composed a long-form, comprehensive, and grounded answers.\footnote{Annotation guidelines and terms are presented in Appendix \ref{sec:guidelines}.}

\paragraph{Question annotation.} We first filter out questions that are either invalid, a false premise, or not information seeking. A question is considered valid if it is understandable, answerable, and not harmful. Each question is then annotated across four dimensions: Time-sensitivity, Complexity, Popularity, and Category.
Figure~\ref{fig:question_ann_statistics} shows the distribution of question annotations.


\paragraph{Grounding relevancy.}
Each grounding contains a piece of text, the source (Web or private KB) and, if available, its age.\footnote{The age of a grounding is the difference between the dates of the question and the grounding.} In total, 4,752 passages retrieved from private knowledge bases and 30,599 from the Web are annotated. 
For each passage, the annotators first identify whether it is relevant to the topic of the question. If so, it is labeled as one of: \textsc{answer-the-question}, \textsc{related-information}, \textsc{outdated}, or \textsc{unknown}. Figure~\ref{fig:grounding_ans_statistics} show the distribution of grounding sources and relevancy.

\paragraph{Answer annotation.}
The annotator writes a long-form answer using only information from the grounding passages that were labelled as relevant.\footnote{The average answer length is 192 words ($\sigma=73.5$), excluding deflections.} 
Our guidelines provide some freedom for the annotators to include \textsc{outdated} or \textsc{unknown} information depending on the time-sensitivity of the question, as long as they are annotated as relevant. 
Annotators extract claims from questions and add citation markers to ground each answer claim in the supporting evidence.\footnote{There are on average 5 unique cited sources per answer.} 
For questions lacking sufficient information for generating a response, annotators write a deflection response, e.g. "There is not enough grounding for an answer.".
A total of 427 questions are labelled as requiring a deflective response, due to the irrelevant nature of the grounding. 
This pattern is more frequent on Fast-Changing questions where it is more likely for the grounding to be outdated, and on questions requiring information from private sources -- where the increased topic specificity leads to a more difficult retrieval setting.
Finally, to ensure high quality of the answers in the benchmark, we conduct an independent human annotation workflow on a sample (N=300) which found every answer to be grounded in at least one grounding passage. Results show that the benchmark is truly grounded in the provided information (out of 2340 total claims, 97\% were grounded in at least 1 chunk).

\section{Evaluation framework}

Thanks to the fine-grained annotations of GaRAGe at the question, grounding, and answer level, we have the opportunity to evaluate several dimensions of model performance. To this end, we follow \citet{jacovi2024facts} and report response eligibility and factuality. We build on top of the factuality metric, by defining metrics reflecting factuality to relevant passages. Then, we evaluate \textit{deflective} behaviour as opposed to hallucination when no relevant grounding is provided, and \textit{attribution} ability to each of the provided relevant passages when generating the final response. 
We compute the metrics using GPT-4o\footnote{\texttt{gpt-4o-2024-11-20} with a temperature of 0.2.} as a judge, unless otherwise specified. 

\paragraph{Eligibility score} measures whether the model sufficiently addresses the user request with respect to the human-written answer as a baseline response. 
The LLM judge produces the labels: No Issues, Minor Issue(s) or Major Issue(s) and the eligibility score is calculated as the percentage of datapoints having no Major Issue(s). The prompt used for this metric is reported in Appendix \ref{app:eligibility_prompt}. 

\paragraph{Unadjusted Factuality score} measures the percentage of answers that are completely supported by the provided grounding, by first splitting the model response into sentences and then judging whether each of those is supported, unsupported, contradictory, or not requiring factual attribution with respect to the provided grounding. Finally, a model answer is factual only if every sentence is supported or does not  require attribution. The prompt used for this metric is reported in Appendix \ref{app:grounding_prompt}. As GaRAGe provides relevancy annotations for each grounding piece, we sharpen this metric to calculate the model's ability to ground its answers only in the relevant passages provided as context, hereinafter \textbf{Unadjusted Relevance-Aware Factuality score (uRAF)}. Notice that this is a more controlled metric that not only measures the factuality of the response, but also the relevancy of the information used to produce the final response. 

 \begin{table*}[]
 \label{tab:main_results}
 \centering
 \small
 \renewcommand{\arraystretch}{1}
 \begin{tabular}{lcc@{\extracolsep{0.8cm}}ccc}
 \toprule
  \textbf{Model} &\begin{tabular}[c]{@{}l@{}} \textbf{Eligibility}  \end{tabular}& \begin{tabular}[c]{@{}l@{}} \textbf{Unadjusted Factuality} \end{tabular} & \begin{tabular}[c]{@{}l@{}}\textbf{Factuality} \end{tabular} & \begin{tabular}[c]{@{}l@{}}  \textbf{uRAF} \end{tabular} & \begin{tabular}[c]{@{}l@{}}\textbf{RAF} \end{tabular} \\ \hline
  \multicolumn{6}{c}{\textit{API-based models}} \\ \hline 
 Claude Haiku  & 79.37 & 51.83 & 48.37 &  40.00  & 36.90 \\ 
 Claude Sonnet & 86.07 & 68.48 & 64.67 & 51.75 & 48.91\\
 Gemini 1.5 & 84.88 & \textbf{81.07} & \textbf{70.50} & 67.78 & 59.43 \\
 Gpt-4o & \textbf{92.47} & 62.91 & 59.30 & 56.18 & 52.88 \\
 Nova Micro & 90.97 & 48.83 & 45.02 & 39.94 & 37.16 \\
 Nova Lite & 80.15 & 61.06 & 49.25 &  55.67 & 45.97  \\
 Nova Pro & 87.77 & 74.47 & 66.63 & \textbf{68.29} & \textbf{60.67} \\ \hline
 \multicolumn{6}{c}{\textit{Open-weights models}} \\ \hline
 Mistral & 85.30 & 48.01 & 43.32  & 38.14 & 34.32 \\
 Mixtral & 82.72 & 47.08 & 42.96 & 37.31 & 34.12 \\
 Qwen 14b & 85.80 & \textit{68.20} & 59.80 & \textit{59.20} & 52.70 \\
 Qwen 32b & \textit{90.50} & 66.10 & \textit{61.00} & 57.70 & \textit{52.90} \\ \bottomrule
 \end{tabular}
 \caption{Performance of different open and closed models on \garage{} measured using LLM-as-a-judge metrics. While Unadjusted Relevance-Aware Factuality Score (\textbf{uRAF}) measures the factuality of the response in relevant information, Relevance-Aware Factuality Score (\textbf{RAF}) measures the percentage of answers that are both eligible and supported by the relevant grounding.}
 \label{tab:main_results}

 \end{table*}

\paragraph{Factuality score} combines the first two metrics, and measures the percentage of answers that are both eligible and supported by the provided grounding. We similarly develop the \textbf{Relevance-Aware Factuality score (RAF)} by combining uRAF with Eligibility to calculate the percentage of answers that are both eligible and supported by the relevant passages.

\paragraph{Deflection score} measures the percentage of deflective answers for the subset of the benchmark expecting a deflection, i.e., \textit{true positive}, due to insufficient grounding information. 
Additionally, we report the portion of \textit{false positive} deflective answers on the rest of the dataset which has sufficient relevant information for providing a non-deflective response. The prompt used for this metric is reported in Appendix \ref{app:deflection_prompt}. 

\paragraph{Attribution score} measures the ability of the model to produce correct attribution to grounding passages from where the information was taken in the form of citations. For this metric, we leverage the citations in the human-written responses and calculate \textit{Precision}, \textit{Recall} and \textit{F$_{1}$-score} of the occurrence of citations in model response compared to those in the human-written response.

\section{Results}
\label{sec:results}

We evaluate strong proprietary models and leading open weights models of different sizes on the GaRAGe benchmark. 
In particular, we evaluate GPT-4o\footnote{\texttt{gpt-4o-2024-11-20}} \citep{hurst2024gpt}, Gemini 1.5 Flash \citep{team2023gemini}, Claude 3.5 Haiku\footnote{\texttt{anthropic.claude-3-5-haiku-20241022-v1:0}} and Sonnet\footnote{\texttt{anthropic.claude-3-5-sonnet-20241022-v2:0}} \citep{anthropic2024introducing}, Amazon Nova Micro, Lite and Pro \citep{intelligence2024amazon}, and open-weights Mistral, Mixtral \citep{jiang2024mixtral}, Qwen2.5 14b and 32b models \citep{yang2024qwen2}. 
We use greedy decoding and the same input prompt across the models as reported in Appendix \ref{app:prompt_example}. 

\subsection{Main Results}
\label{sec:main_results}

In Table~\ref{tab:main_results} we report the main results in terms of answer eligibility and factuality. Firstly, GPT-4o attains the highest eligibility score.
Surprisingly, the size of the model does not have a clear impact on this metric: both Nova Micro and Mistral outperform bigger models from the same family, while Claude 3.5 Haiku lags behind Sonnet.
Furthermore, Gemini 1.5 Flash achieves the highest factuality scores, followed by Nova Pro and then Claude 3.5 Sonnet and GPT-4o.
The size of the model is an indicator of the model capacity in factuality, suggesting that bigger models are constantly making better use of the provided information as context. 
Likely GPT-4o is generating information from parametric knowledge, which while useful as reported by eligibility score, goes beyond the relevant information provided in the grounding which might indicate a risk of hallucination.
By combining eligibility and unadjusted factuality scores, the factuality score favors eligible answers that make full use of the provided grounding. In this metric, Gemini 1.5 Flash is leading, closely followed by Nova Pro and Claude 3.5 Sonnet.

Interestingly, when instead restricting the grounding score to only the relevant grounding, we notice an overall drop in performance for most of the models. This demonstrates that most of the models are utilizing irrelevant or outdated grounding when generating the response, i.e., acting as a summariser, rather than identifying only the fresh and relevant information to include in the final response.  
The leading models in terms of \textit{RAF score} are Nova Pro and Gemini 1.5 Flash, which achieve $\sim$60\% in both eligible and relevant-information grounded responses. \\

\begin{table}[t!]
    \centering
     \resizebox{.8\columnwidth}{!}{%
    \begin{tabular}{lcc}
\toprule
         \textbf{Model} & \textbf{True Positive} & \textbf{False Positive} \\
         \midrule
         Claude Haiku & 15.2 & 0.4 \\
         Claude Sonnet & 25.3 & 1.4 \\
         Gemini 1.5 & 27.2 & 2.3 \\
         GPT-4o & \textbf{31.1} & 2.3 \\
         Nova Micro & 7.0 & \textit{0.3} \\
         Nova Lite & 7.5 & 1.4 \\
         Nova Pro & 18.0 & 0.8 \\
         Mistral & 5.2 & 0.8 \\
         Mixtral & 9.4 & 3.3 \\
         Qwen 14b & 17.1 & 1.1 \\
         Qwen 32b & 21.5 & 1.2 \\
         \bottomrule
    \end{tabular}
    }
    \caption{True positive and False Positive deflective responses on the respective subsets of the benchmark.}
    \label{tab:deflection}

\end{table}
\noindent
In Table~\ref{tab:deflection} we report the average deflective answers on the deflection subset as true positives, and false positives on the subset with sufficient information to provide a non-deflective response.
Firstly, all the baseline models have a low false positive rate (less than  3.5\%), meaning that when sufficient relevant information is provided in the grounding passages, the models correctly provide a non-deflective answer. 
On the other hand, the models struggle at identifying the lack of sufficient grounding for generating a good answer.
While larger models show a better deflective behaviour overall, the deflection coverage as measured by \textit{true positive} rate is rather low, with the best performing model being GPT-4o with 31.1\%. This is a significant gap for all models to improve on, to avoid hallucinations and misinforming RAG users. \\

\noindent
In Table~\ref{tab:attribution} we report the attribution score in terms of Precision, Recall and F$_{1}$-score.
Results show that larger models exhibit a more conservative approach towards attribution, having lower values on recall, but compensating through higher numbers for precision. 
This indicates that smaller models might be over-citing the provided sources rather than focusing only on the relevant grounding.
Overall, we observe a similar performance across most of the models, ranging between 50-60\% in F$_{1}$-score indicating a gap of the models to attribute the generated response to the corresponding grounding information.

\subsection{Analysis}
 \begin{figure*}[t]
    \centering
    \includegraphics[width=1\textwidth]{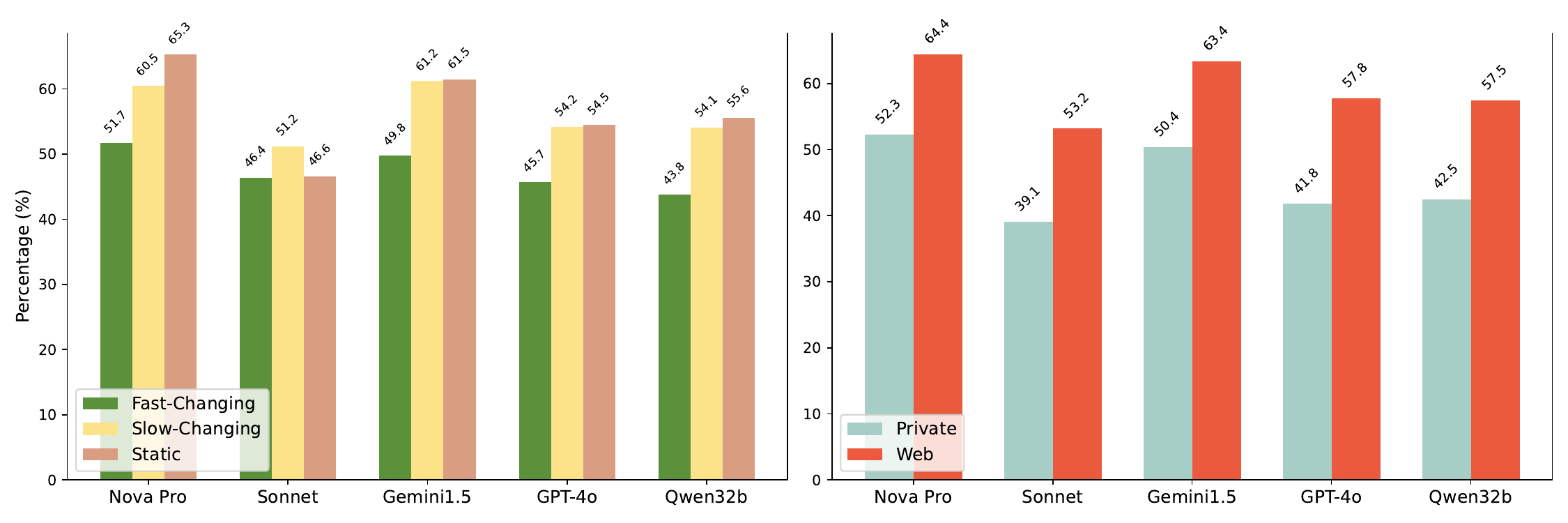}
    \caption{Performance of different models measured by RAF score on 1) different temporal question dynamism on the left and 2) private versus Web questions on the right.
    }
    \label{fig:ent_Web_results}
\end{figure*}

\paragraph{Time-sensitivity.} We analyze the performance on the \textit{fast-changing} questions of the benchmark with the assumption that (1) grounding passages are likely noisier and outdated, (2) LLMs struggle to reason about the relative recency of the question and grounding information. In the first plot in Figure ~\ref{fig:ent_Web_results} we report RAF scores for the bigger models. Results show a similar trend to the main results (\S\ref{sec:main_results}), with Nova Pro achieving a score of 51.7\%, followed by Gemini (49.8\%), Sonnet (46.7\%), GPT-4o (45.7\%) and Qwen 32b (43.8\%). Interestingly, the performance on this set of questions is generally lower ($\sim$10\% across models) with respect to the slow-changing or static questions, demonstrating that LLMs struggle more when reasoning over temporal facts. 

\begin{table}[t!]
    \centering
    \resizebox{.8\columnwidth}{!}{%
    \begin{tabular}{lc@{\extracolsep{0.2cm}}c@{\extracolsep{0.5cm}}c}
\toprule
         \textbf{Model} & \textbf{Precision}  &  \textbf{Recall} & \textbf{F$_{1}$} \\ \midrule
         Claude Haiku & 49.9 & 71.9 & \textit{58.9}\\
         Claude Sonnet & 51.8 & 67.5 & 58.6 \\
         Gemini 1.5 & 54.7 & 56.3 & 55.5 \\
         GPT-4o & \textbf{57.9} & 59.0 & 58.4 \\
         Nova Micro & 48.2 & \textbf{75.8}  & \textit{58.9} \\
         Nova Lite  & 50.0 & 46.4  & 48.1\\
         Nova Pro & 56.9 & 49.6  & 53.0 \\
         Mistral & 44.4 & 60.0  & 51.0 \\
         Mixtral & 46.3 & 56.3 & 51.0 \\
         Qwen 14b & 55.2 & 43.6 & 48.7\\
         Qwen 32b & 55.2 & 47.0 & 50.8 \\
         \bottomrule
    \end{tabular}
    }
    \caption{Performance of the models measured by Attribution Precision, Recall and F$_{1}$ in GaRAGe.}
    \label{tab:attribution}
\end{table}

\paragraph{Private questions performance.}

We further analyze the quality of the grounding for questions built from specific domains that simulate private knowledge bases in comparison with questions reflecting general Web searches.
The domain questions have less relevant grounding in comparison with the Web ones, likely making them more challenging for LLMs to extract only the relevant information. For example, the questions covering Enron topics have 47.8\% relevant grounding on average while questions with more broad topics reflecting Web usage have 85.6\% relevant passages.

Results in Figure \ref{fig:ent_Web_results} show that there is a significant decrease in performance even across the strongest evaluated LLMs with a constant drop of more than 10\% for the questions addressing more specific topics. 
This confirms our intuition and further suggests that LLMs struggle to create grounded answers to the relevant information only. 

\paragraph{Grounding quality.} We investigate how the grounding relevancy influences the answer quality, in order to understand the model's robustness to irrelevant information. We divide the benchmark into three different groups using the percentage of relevant passages ($pass_r$) provided as gounding: low ($ pass_r <33\%$), medium ($33\% <= pass_r  < 66\%$) and high ( $pass_r >66\%$).  

\begin{figure}[t]
    \centering
    \includegraphics[width=1\columnwidth]{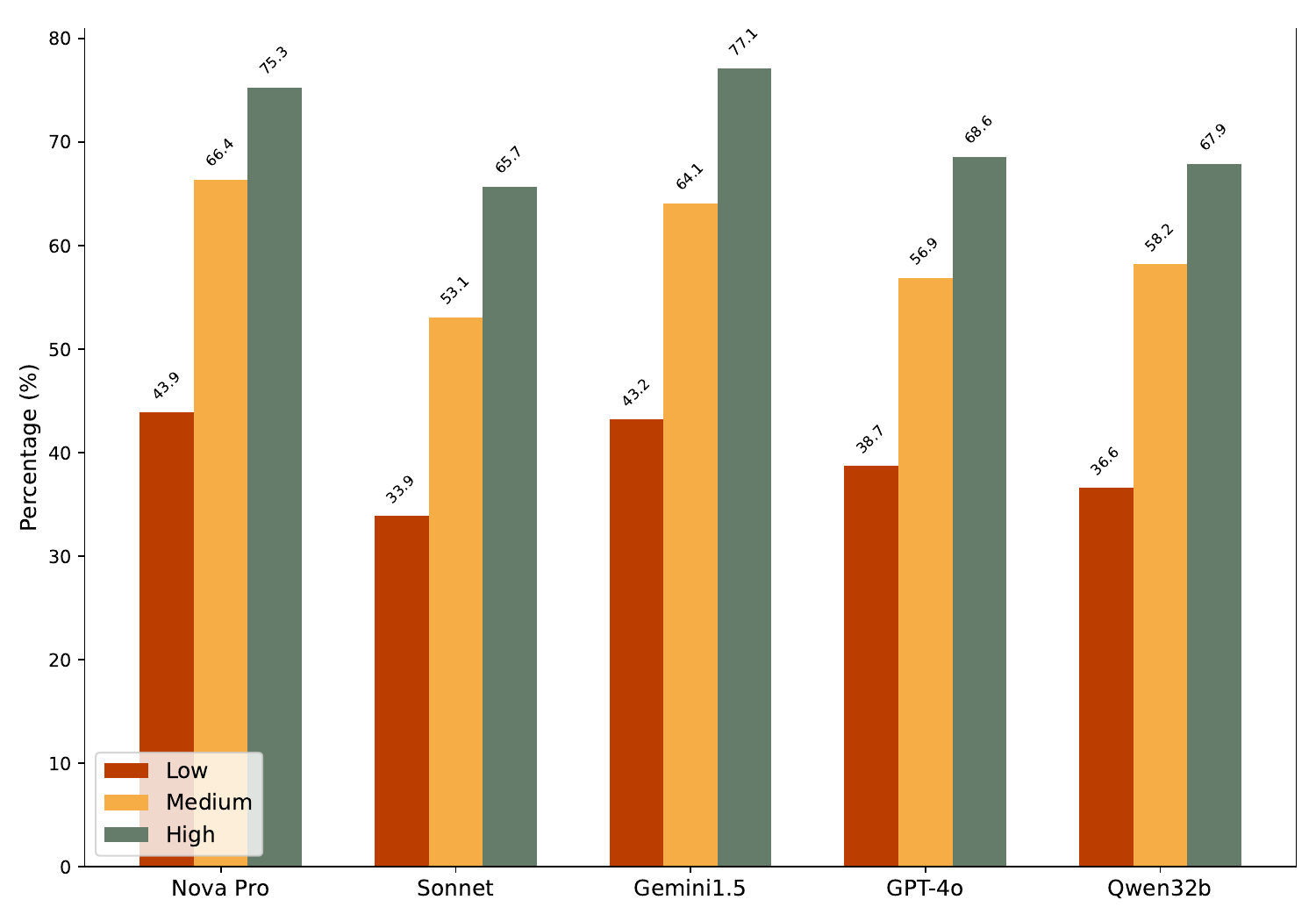}
    \caption{Performance of different models measured by RAF score relative to the grounding quality.
    }
    \label{fig:grounding_quality}
\end{figure}

Figure \ref{fig:grounding_quality} presents the RAF score of the bigger models we report in our benchmark. We observe a clear trend with the models performing significantly worse as the grounding quality degrades with a 10\% drop in performance from high to medium and with 20\% drop from medium to low. This pattern highlights the incapability of the models to distinguish between relevant and irrelevant information when building an answer in the RAG setup.

\paragraph{Question popularity.}

When analyzing the performance of the models according to question popularity, we find that the RAF scores for Tail questions is 44.2\% when we average across models. This is significantly lower to both Torso and Head questions which have an average RAF score of 49.7\% and 50.3\%, respectively.   
These results exhibit similar patterns with the ones presented in the previous analysis, where Tail questions generally are more challenging due to the scarcity of relevant information on those topics and potentially noiser grounding.

\section{Related Work}
There have been an impressive number of benchmarks released in the last years in the domain of Question Answering, due to both the impact of the task and the complex and diverse reasoning required to have in order to solve it, spawning from extractive benchmark probing for reading comprehension like: Natural Questions \citep{kwiatkowski2019natural}, WikiQA \citep{yang2015wikiqa} or TriviaQA \citep{joshi2017triviaqa} to frameworks that are easily customizable to new data and models such as: Ragas \citep{es2024ragas}, Ragnarok \citep{pradeep2024ragnar} or Ares \citep{saad2024ares}.

\paragraph{Question complexity.} A dimension which evolved lately for challenging LLM capabilites on RAG was to increase the complexity of the questions asked. This was done either by requiring a more complex reasoning over multiple documents \citep{tang2024multihoprag, trivedi-etal-2022-musique} or understanding extra-linguistic contextual information like temporal information in order to construct the correct answer \citep{kasai2024realtime, zhang2021situatedqa}. \citet{gabburo-etal-2024-measuring} propose a metric for measuring question complexity based on the retrieved evidences, capturing the difficulty of finding accurate answers from multiple grounding sources.

\paragraph{Grounding Coverage.} Another component that can be difficult for LLMs to deal with is the variety of the provided grounding to the model. This ranges from grounding with different levels of noise \citep{chen2024benchmarking, liu2023recall}, specific domains \citep{guinet2024automated, han2023robustqa} or combining grounding from both private and public sources \citep{arora2023reasoning}.

\paragraph{Answer Comprehesiveness.} Most of the RAG benchmarks focus on very short answers. This pattern is due to both the complexity of creating long, comprehensive and grounded answers, but also due to the difficulty to evaluate them. In the same time, LLMs generally have the default behaviour to be verbose with their answers, so there is a mismatch between the model behaviour and the evaluation frameworks for RAG usecases. This is mitigated lately through efforts of evaluating long-form Q\&A with benchmarks like: Rag-QA Arena\citep{han2024rag}, Facts \citep{jacovi2024facts}, LongFact \citep{wei2024long} or ELI5 \citep{fan2019eli5}.

\section{Conclusion}
We introduce GaRAGe, a large RAG benchmark with human-curated long-form answers and comprehensive annotations of individual grounding passages. Due to the detailed annotation of the dataset, we were able to develop a thorough set of experiments and a new metric: Reference-Aware Factuality (RAF) Score. 
This framework helped unveil multiple gaps in open and closed LLM's abilities to deal with noisy grounding. Furthermore, we conduct in-depth analysis on model performance across different dataset dimensions, increasing our understanding of model limitations.

\newpage

\section*{Limitations}
\label{sec:limitations}
We acknowledge that the benchmark contains only English datapoints. We plan to explore multilingual settings in the future.

The dataset might be included in future LLM training which would invalidate the evaluation on the benchmark to some degree. Even though the data might be included in the future (or some portions are already included), the design of the dataset which measures the capability of the model to follow the provided grounding ensures the validity of the benchmark.

Some components of the annotations are somewhat subjective, for example the popularity of a topic can be different between annotators. While this can create some misaligned labels, we consider the significant size of the dataset alleviates these outliers. 

Due to the fact that a significant part of the benchmark reflects realistic internet questions, they might refer to individual people. In order to ensure the absence of offensive content, the first filter for the question annotation is that the question should not request sensitive information; nor contain offensive, abusive, or harmful language. It should not be discriminative based on gender or race.

The LLMs might have different behaviour when we run them multiple times. Due to the costs for running the model inference, we report the performance on one utterance and we acknowledge that there might be some fluctuations along multiple runs.

We use only GPT-4o as a judge, so the results might be biased towards GPT models during evaluation \citep{liu2024llms}. This can be mitigated either through running multiple LLM judges and aggregate their judging for a more robust report, or through human evaluation.

\section*{Ethics statement}
\label{sec:ethics}
We utilize publicly available news data, which may contain viewpoints from different perspectives and private knoweldge bases that were previously used in publications. The output results in the paper do not represent the views of the authors.

\newpage

\label{sec:bibtex}

\bibliography{acl_latex}

\begin{thebibliography}{39}
\providecommand{\natexlab}[1]{#1}

\bibitem[{Anthropic(2024)}]{anthropic2024introducing}
AI~Anthropic. 2024.
\newblock Introducing the next generation of claude.

\bibitem[{Arora et~al.(2023)Arora, Lewis, Fan, Kahn, and R{\'e}}]{arora2023reasoning}
Simran Arora, Patrick Lewis, Angela Fan, Jacob Kahn, and Christopher R{\'e}. 2023.
\newblock Reasoning over public and private data in retrieval-based systems.
\newblock \emph{Transactions of the Association for Computational Linguistics}, 11:902--921.

\bibitem[{Chen et~al.(2024{\natexlab{a}})Chen, Xiao, Zhang, Luo, Lian, and Liu}]{chen2024bge}
Jianlv Chen, Shitao Xiao, Peitian Zhang, Kun Luo, Defu Lian, and Zheng Liu. 2024{\natexlab{a}}.
\newblock \href {https://arxiv.org/abs/2402.03216} {Bge m3-embedding: Multi-lingual, multi-functionality, multi-granularity text embeddings through self-knowledge distillation}.
\newblock \emph{Preprint}, arXiv:2402.03216.

\bibitem[{Chen et~al.(2024{\natexlab{b}})Chen, Lin, Han, and Sun}]{chen2024benchmarking}
Jiawei Chen, Hongyu Lin, Xianpei Han, and Le~Sun. 2024{\natexlab{b}}.
\newblock Benchmarking large language models in retrieval-augmented generation.
\newblock In \emph{Proceedings of the AAAI Conference on Artificial Intelligence}, volume~38, pages 17754--17762.

\bibitem[{Es et~al.(2024)Es, James, Anke, and Schockaert}]{es2024ragas}
Shahul Es, Jithin James, Luis~Espinosa Anke, and Steven Schockaert. 2024.
\newblock Ragas: Automated evaluation of retrieval augmented generation.
\newblock In \emph{Proceedings of the 18th Conference of the European Chapter of the Association for Computational Linguistics: System Demonstrations}, pages 150--158.

\bibitem[{Fan et~al.(2019)Fan, Jernite, Perez, Grangier, Weston, and Auli}]{fan2019eli5}
Angela Fan, Yacine Jernite, Ethan Perez, David Grangier, Jason Weston, and Michael Auli. 2019.
\newblock Eli5: Long form question answering.
\newblock In \emph{Proceedings of the 57th Annual Meeting of the Association for Computational Linguistics}, pages 3558--3567.

\bibitem[{Gabburo et~al.(2024)Gabburo, Jedema, Garg, Ribeiro, and Moschitti}]{gabburo-etal-2024-measuring}
Matteo Gabburo, Nicolaas~Paul Jedema, Siddhant Garg, Leonardo F.~R. Ribeiro, and Alessandro Moschitti. 2024.
\newblock \href {https://doi.org/10.18653/v1/2024.findings-acl.872} {Measuring retrieval complexity in question answering systems}.
\newblock In \emph{Findings of the Association for Computational Linguistics: ACL 2024}, pages 14636--14650, Bangkok, Thailand. Association for Computational Linguistics.

\bibitem[{Guinet et~al.(2024)Guinet, Omidvar-Tehrani, Deoras, and Callot}]{guinet2024automated}
Gauthier Guinet, Behrooz Omidvar-Tehrani, Anoop Deoras, and Laurent Callot. 2024.
\newblock Automated evaluation of retrieval-augmented language models with task-specific exam generation.
\newblock \emph{arXiv preprint arXiv:2405.13622}.

\bibitem[{Han et~al.(2023)Han, Qi, Zhang, Liu, Burger, Wang, Huang, Xiang, and Roth}]{han2023robustqa}
Rujun Han, Peng Qi, Yuhao Zhang, Lan Liu, Juliette Burger, William~Yang Wang, Zhiheng Huang, Bing Xiang, and Dan Roth. 2023.
\newblock Robustqa: Benchmarking the robustness of domain adaptation for open-domain question answering.
\newblock In \emph{Findings of the Association for Computational Linguistics: ACL 2023}, pages 4294--4311.

\bibitem[{Han et~al.(2024)Han, Zhang, Qi, Xu, Wang, Liu, Wang, Min, and Castelli}]{han2024rag}
Rujun Han, Yuhao Zhang, Peng Qi, Yumo Xu, Jenyuan Wang, Lan Liu, William~Yang Wang, Bonan Min, and Vittorio Castelli. 2024.
\newblock Rag-qa arena: Evaluating domain robustness for long-form retrieval augmented question answering.
\newblock \emph{arXiv preprint arXiv:2407.13998}.

\bibitem[{Hurst et~al.(2024)Hurst, Lerer, Goucher, Perelman, Ramesh, Clark, Ostrow, Welihinda, Hayes, Radford et~al.}]{hurst2024gpt}
Aaron Hurst, Adam Lerer, Adam~P Goucher, Adam Perelman, Aditya Ramesh, Aidan Clark, AJ~Ostrow, Akila Welihinda, Alan Hayes, Alec Radford, et~al. 2024.
\newblock Gpt-4o system card.
\newblock \emph{arXiv preprint arXiv:2410.21276}.

\bibitem[{Intelligence(2024)}]{intelligence2024amazon}
Amazon Artificial~General Intelligence. 2024.
\newblock The amazon nova family of models: Technical report and model card.

\bibitem[{Jacovi et~al.(2024)Jacovi, Wang, Alberti, Tao, Lipovetz, Olszewska, Haas, Liu, Keating, Bloniarz, Saroufim, Fry, Marcus, Kukliansky, Tomar, Swirhun, Xing, Wang, Gurumurthy, Aaron, Ambar, Fellinger, Wang, Zhang, Goldshtein, and Das}]{jacovi2024facts}
Alon Jacovi, Andrew Wang, Chris Alberti, Connie Tao, Jon Lipovetz, Kate Olszewska, Lukas Haas, Michelle Liu, Nate Keating, Adam Bloniarz, Carl Saroufim, Corey Fry, Dror Marcus, Doron Kukliansky, Gaurav~Singh Tomar, James Swirhun, Jinwei Xing, Lily Wang, Madhu Gurumurthy, Michael Aaron, Moran Ambar, Rachana Fellinger, Rui Wang, Zizhao Zhang, Sasha Goldshtein, and Dipanjan Das. 2024.
\newblock {FACTS} grounding: A new benchmark for evaluating the factuality of large language models.

\bibitem[{Jiang et~al.(2024)Jiang, Sablayrolles, Roux, Mensch, Savary, Bamford, Chaplot, Casas, Hanna, Bressand et~al.}]{jiang2024mixtral}
Albert~Q Jiang, Alexandre Sablayrolles, Antoine Roux, Arthur Mensch, Blanche Savary, Chris Bamford, Devendra~Singh Chaplot, Diego de~las Casas, Emma~Bou Hanna, Florian Bressand, et~al. 2024.
\newblock Mixtral of experts.
\newblock \emph{arXiv preprint arXiv:2401.04088}.

\bibitem[{Joshi et~al.(2017)Joshi, Choi, Weld, and Zettlemoyer}]{joshi2017triviaqa}
Mandar Joshi, Eunsol Choi, Daniel~S Weld, and Luke Zettlemoyer. 2017.
\newblock Triviaqa: A large scale distantly supervised challenge dataset for reading comprehension.
\newblock \emph{arXiv preprint arXiv:1705.03551}.

\bibitem[{Kasai et~al.(2024)Kasai, Sakaguchi, Le~Bras, Asai, Yu, Radev, Smith, Choi, Inui et~al.}]{kasai2024realtime}
Jungo Kasai, Keisuke Sakaguchi, Ronan Le~Bras, Akari Asai, Xinyan Yu, Dragomir Radev, Noah~A Smith, Yejin Choi, Kentaro Inui, et~al. 2024.
\newblock Realtime qa: what's the answer right now?
\newblock \emph{Advances in Neural Information Processing Systems}, 36.

\bibitem[{Klimt and Yang(2004)}]{klimt2004introducing}
Bryan Klimt and Yiming Yang. 2004.
\newblock Introducing the enron corpus.
\newblock In \emph{CEAS}, volume~4, page~1.

\bibitem[{Kwiatkowski et~al.(2019)Kwiatkowski, Palomaki, Redfield, Collins, Parikh, Alberti, Epstein, Polosukhin, Devlin, Lee et~al.}]{kwiatkowski2019natural}
Tom Kwiatkowski, Jennimaria Palomaki, Olivia Redfield, Michael Collins, Ankur Parikh, Chris Alberti, Danielle Epstein, Illia Polosukhin, Jacob Devlin, Kenton Lee, et~al. 2019.
\newblock Natural questions: a benchmark for question answering research.
\newblock \emph{Transactions of the Association for Computational Linguistics}, 7:453--466.

\bibitem[{Laban et~al.(2024)Laban, Fabbri, Xiong, and Wu}]{laban-etal-2024-summary}
Philippe Laban, Alexander Fabbri, Caiming Xiong, and Chien-Sheng Wu. 2024.
\newblock \href {https://doi.org/10.18653/v1/2024.emnlp-main.552} {Summary of a haystack: A challenge to long-context {LLM}s and {RAG} systems}.
\newblock In \emph{Proceedings of the 2024 Conference on Empirical Methods in Natural Language Processing}, pages 9885--9903, Miami, Florida, USA. Association for Computational Linguistics.

\bibitem[{Lewis et~al.(2020)Lewis, Perez, Piktus, Petroni, Karpukhin, Goyal, K\"{u}ttler, Lewis, Yih, Rockt\"{a}schel, Riedel, and Kiela}]{NEURIPS2020_6b493230}
Patrick Lewis, Ethan Perez, Aleksandra Piktus, Fabio Petroni, Vladimir Karpukhin, Naman Goyal, Heinrich K\"{u}ttler, Mike Lewis, Wen-tau Yih, Tim Rockt\"{a}schel, Sebastian Riedel, and Douwe Kiela. 2020.
\newblock \href {https://proceedings.neurips.cc/paper_files/paper/2020/file/6b493230205f780e1bc26945df7481e5-Paper.pdf} {Retrieval-augmented generation for knowledge-intensive nlp tasks}.
\newblock In \emph{Advances in Neural Information Processing Systems}, volume~33, pages 9459--9474. Curran Associates, Inc.

\bibitem[{Liu et~al.(2023)Liu, Huang, Li, Chen, Zhou, Meng, Zhou, and Sun}]{liu2023recall}
Yi~Liu, Lianzhe Huang, Shicheng Li, Sishuo Chen, Hao Zhou, Fandong Meng, Jie Zhou, and Xu~Sun. 2023.
\newblock Recall: A benchmark for llms robustness against external counterfactual knowledge.
\newblock \emph{arXiv preprint arXiv:2311.08147}.

\bibitem[{Liu et~al.(2024)Liu, Moosavi, and Lin}]{liu2024llms}
Yiqi Liu, Nafise Moosavi, and Chenghua Lin. 2024.
\newblock Llms as narcissistic evaluators: When ego inflates evaluation scores.
\newblock In \emph{The 62nd Annual Meeting of the Association for Computational Linguistics}.

\bibitem[{Luo et~al.(2025)Luo, E, Guo, Lin, Wu, Mu, Liu, Song, Zhu, and Tuan}]{luo2025kbqao1agenticknowledgebase}
Haoran Luo, Haihong E, Yikai Guo, Qika Lin, Xiaobao Wu, Xinyu Mu, Wenhao Liu, Meina Song, Yifan Zhu, and Luu~Anh Tuan. 2025.
\newblock \href {https://arxiv.org/abs/2501.18922} {Kbqa-o1: Agentic knowledge base question answering with monte carlo tree search}.
\newblock \emph{Preprint}, arXiv:2501.18922.

\bibitem[{Niu et~al.(2024)Niu, Wu, Zhu, Xu, Shum, Zhong, Song, and Zhang}]{niu-etal-2024-ragtruth}
Cheng Niu, Yuanhao Wu, Juno Zhu, Siliang Xu, KaShun Shum, Randy Zhong, Juntong Song, and Tong Zhang. 2024.
\newblock \href {https://doi.org/10.18653/v1/2024.acl-long.585} {{RAGT}ruth: A hallucination corpus for developing trustworthy retrieval-augmented language models}.
\newblock In \emph{Proceedings of the 62nd Annual Meeting of the Association for Computational Linguistics (Volume 1: Long Papers)}, pages 10862--10878, Bangkok, Thailand. Association for Computational Linguistics.

\bibitem[{Oh and Thorne(2023)}]{oh-thorne-2023-detrimental}
Philhoon Oh and James Thorne. 2023.
\newblock \href {https://doi.org/10.18653/v1/2023.findings-emnlp.776} {Detrimental contexts in open-domain question answering}.
\newblock In \emph{Findings of the Association for Computational Linguistics: EMNLP 2023}, pages 11589--11605, Singapore. Association for Computational Linguistics.

\bibitem[{Pradeep et~al.(2024)Pradeep, Thakur, Sharifymoghaddam, Zhang, Nguyen, Campos, Craswell, and Lin}]{pradeep2024ragnar}
Ronak Pradeep, Nandan Thakur, Sahel Sharifymoghaddam, Eric Zhang, Ryan Nguyen, Daniel Campos, Nick Craswell, and Jimmy Lin. 2024.
\newblock Ragnar$\backslash$" ok: A reusable rag framework and baselines for trec 2024 retrieval-augmented generation track.
\newblock \emph{arXiv preprint arXiv:2406.16828}.

\bibitem[{Reimers and Gurevych(2019)}]{reimers-gurevych-2019-sentence}
Nils Reimers and Iryna Gurevych. 2019.
\newblock \href {https://doi.org/10.18653/v1/D19-1410} {Sentence-{BERT}: Sentence embeddings using {S}iamese {BERT}-networks}.
\newblock In \emph{Proceedings of the 2019 Conference on Empirical Methods in Natural Language Processing and the 9th International Joint Conference on Natural Language Processing (EMNLP-IJCNLP)}, pages 3982--3992, Hong Kong, China. Association for Computational Linguistics.

\bibitem[{Saad-Falcon et~al.(2024)Saad-Falcon, Khattab, Potts, and Zaharia}]{saad2024ares}
Jon Saad-Falcon, Omar Khattab, Christopher Potts, and Matei Zaharia. 2024.
\newblock Ares: An automated evaluation framework for retrieval-augmented generation systems.
\newblock In \emph{Proceedings of the 2024 Conference of the North American Chapter of the Association for Computational Linguistics: Human Language Technologies (Volume 1: Long Papers)}, pages 338--354.

\bibitem[{Tang and Yang(2024)}]{tang2024multihoprag}
Yixuan Tang and Yi~Yang. 2024.
\newblock \href {https://openreview.net/forum?id=t4eB3zYWBK} {Multihop-{RAG}: Benchmarking retrieval-augmented generation for multi-hop queries}.
\newblock In \emph{First Conference on Language Modeling}.

\bibitem[{Team et~al.(2023)Team, Anil, Borgeaud, Alayrac, Yu, Soricut, Schalkwyk, Dai, Hauth, Millican et~al.}]{team2023gemini}
Gemini Team, Rohan Anil, Sebastian Borgeaud, Jean-Baptiste Alayrac, Jiahui Yu, Radu Soricut, Johan Schalkwyk, Andrew~M Dai, Anja Hauth, Katie Millican, et~al. 2023.
\newblock Gemini: a family of highly capable multimodal models.
\newblock \emph{arXiv preprint arXiv:2312.11805}.

\bibitem[{Trivedi et~al.(2022)Trivedi, Balasubramanian, Khot, and Sabharwal}]{trivedi-etal-2022-musique}
Harsh Trivedi, Niranjan Balasubramanian, Tushar Khot, and Ashish Sabharwal. 2022.
\newblock \href {https://doi.org/10.1162/tacl_a_00475} {{M}u{S}i{Q}ue: Multihop questions via single-hop question composition}.
\newblock \emph{Transactions of the Association for Computational Linguistics}, 10:539--554.

\bibitem[{Vodrahalli et~al.(2024)Vodrahalli, Ontanon, Tripuraneni, Xu, Jain, Shivanna, Hui, Dikkala, Kazemi, Fatemi et~al.}]{vodrahalli2024michelangelo}
Kiran Vodrahalli, Santiago Ontanon, Nilesh Tripuraneni, Kelvin Xu, Sanil Jain, Rakesh Shivanna, Jeffrey Hui, Nishanth Dikkala, Mehran Kazemi, Bahare Fatemi, et~al. 2024.
\newblock Michelangelo: Long context evaluations beyond haystacks via latent structure queries.
\newblock \emph{arXiv preprint arXiv:2409.12640}.

\bibitem[{Wei et~al.(2024{\natexlab{a}})Wei, Karina, Chung, Jiao, Papay, Glaese, Schulman, and Fedus}]{wei2024measuring}
Jason Wei, Nguyen Karina, Hyung~Won Chung, Yunxin~Joy Jiao, Spencer Papay, Amelia Glaese, John Schulman, and William Fedus. 2024{\natexlab{a}}.
\newblock Measuring short-form factuality in large language models.
\newblock \emph{arXiv preprint arXiv:2411.04368}.

\bibitem[{Wei et~al.(2024{\natexlab{b}})Wei, Yang, Song, Lu, Hu, Tran, Peng, Liu, Huang, Du et~al.}]{wei2024long}
Jerry Wei, Chengrun Yang, Xinying Song, Yifeng Lu, Nathan Hu, Dustin Tran, Daiyi Peng, Ruibo Liu, Da~Huang, Cosmo Du, et~al. 2024{\natexlab{b}}.
\newblock Long-form factuality in large language models.
\newblock \emph{arXiv preprint arXiv:2403.18802}.

\bibitem[{Yang et~al.(2024{\natexlab{a}})Yang, Yang, Zhang, Hui, Zheng, Yu, Li, Liu, Huang, Wei et~al.}]{yang2024qwen2}
An~Yang, Baosong Yang, Beichen Zhang, Binyuan Hui, Bo~Zheng, Bowen Yu, Chengyuan Li, Dayiheng Liu, Fei Huang, Haoran Wei, et~al. 2024{\natexlab{a}}.
\newblock Qwen2. 5 technical report.
\newblock \emph{arXiv preprint arXiv:2412.15115}.

\bibitem[{Yang et~al.(2024{\natexlab{b}})Yang, Sun, Xin, Sun, Bhalla, Chen, Choudhary, Gui, Jiang, Jiang, Kong, Moran, Wang, Xu, Yan, Yang, Yuan, Zha, Tang, Chen, Scheffer, Liu, Shah, Wanga, Kumar, tau Yih, and Dong}]{yang2024cragcomprehensiverag}
Xiao Yang, Kai Sun, Hao Xin, Yushi Sun, Nikita Bhalla, Xiangsen Chen, Sajal Choudhary, Rongze~Daniel Gui, Ziran~Will Jiang, Ziyu Jiang, Lingkun Kong, Brian Moran, Jiaqi Wang, Yifan~Ethan Xu, An~Yan, Chenyu Yang, Eting Yuan, Hanwen Zha, Nan Tang, Lei Chen, Nicolas Scheffer, Yue Liu, Nirav Shah, Rakesh Wanga, Anuj Kumar, Wen tau Yih, and Xin~Luna Dong. 2024{\natexlab{b}}.
\newblock \href {https://arxiv.org/abs/2406.04744} {Crag -- comprehensive rag benchmark}.
\newblock \emph{Preprint}, arXiv:2406.04744.

\bibitem[{Yang et~al.(2024{\natexlab{c}})Yang, Sun, Xin, Sun, Bhalla, Chen, Choudhary, Gui, Jiang, Jiang et~al.}]{yang2024crag}
Xiao Yang, Kai Sun, Hao Xin, Yushi Sun, Nikita Bhalla, Xiangsen Chen, Sajal Choudhary, Rongze~Daniel Gui, Ziran~Will Jiang, Ziyu Jiang, et~al. 2024{\natexlab{c}}.
\newblock Crag--comprehensive rag benchmark.
\newblock \emph{arXiv preprint arXiv:2406.04744}.

\bibitem[{Yang et~al.(2015)Yang, Yih, and Meek}]{yang2015wikiqa}
Yi~Yang, Wen-tau Yih, and Christopher Meek. 2015.
\newblock Wikiqa: A challenge dataset for open-domain question answering.
\newblock In \emph{Proceedings of the 2015 conference on empirical methods in natural language processing}, pages 2013--2018.

\bibitem[{Zhang and Choi(2021)}]{zhang2021situatedqa}
Michael~J.Q. Zhang and Eunsol Choi. 2021.
\newblock {S}ituated{QA}: Incorporating extra-linguistic contexts into {QA}.
\newblock \emph{Proceedings of the Conference on Empirical Methods in Natural Language Processing (EMNLP)}.

\end{thebibliography}

\clearpage
\newpage

\appendix

\onecolumn

\section{Annotation definitions}
\label{sec:guidelines}

We present below the definitions used for the annotation guidelines for all the steps.

\subsection{Question dimensions}
\subsubsection{Question filters}

\hspace{\parindent} \textbf{Question validity.} A question is valid if it can be answered by seeking information in the Web, or using other sources of information or common knowledge. Also, it should not request sensitive information; nor contain offensive, abusive, or harmful language. It should not be discriminative based on gender or race.

\textbf{False premise.} False premise questions are questions that contain falsely assumed facts that are not directly stated but are likely to be believed by the asker.

\textbf{Information Seeking.} An Info-seeking question refers to a question posed with the aim of acquiring knowledge, clarity or insights on a particular topic, seeking factual information, or explanation on details.

\subsubsection{Temporal dynamism}
\hspace{\parindent} \textbf{Time sensitive questions.} A time sensitive question refers to a question for which the answer changes with the time and it requires real-time content to be answered. If you are not familiar with the topic, you should  search the Internet in order to make a decision whether it is time-sensitive.

\textbf{Static questions.} A static question refers to a question for which the answer doesn't change with the time or a question that doesn't require real-time content to be answered. 

\textbf{Slow-Changing questions.} It is acceptable to have information be more than a week recency to answer the question. This includes more than one month, more than one year, and so on.Note that if there is trending-news about the same question, it should be considered as less than one week.

\textbf{Fast-Changing questions.} Required information needs to be within a seven day  recency to answer the question.

\subsubsection{Question complexity}

\hspace{\parindent}\textbf{Simple.} Questions asking for simple facts

\textbf{Simple w. condition.} Questions asking for simple facts with some given conditions, such as stock prices on a certain date and a director’s recent movies in a certain genre.

\textbf{Set.} Questions that expect a set of entities or objects as the answer.

\textbf{Comparison.} Questions that compare two entities.

\textbf{Aggregation.} Questions that require aggregation of retrieval results to answer.

\textbf{Multi-hop.} Questions that require chaining multiple pieces of information to compose the answer.

\textbf{Post-processing heavy.} Questions that need reasoning or processing of the retrieved information to obtain the answer.

\subsubsection{Question popularity}

\hspace{\parindent}\textbf{Head.} Covers widely-known, frequently discussed subjects, which receive signficant media coverage. It resembles high-frequency search terms or trending topics.

\textbf{Torso.} Covers moderately popular topics, but not mainstream. It resembles medium-frequency search terms.

\textbf{Tail.} asks  about niche or specialized topics and covers highly technical or specialized fields.

\subsubsection{Question category}

\hspace{\parindent}We cover the following question categories: Arts \& Entertainment, Computers \& Electronics, Health, Jobs \& Education, Home \& Garden,	Law \& Government, Travel, Science, Business \& Industrial, Hobbies \& Leisure, Books \& Literature, Sports, News, Beauty \& Fitness, Finance, People \& Society, Autos \& Vehicles,	Games, Time \& Weather, Online Communities, Internet \& Telecom, Local Information,	Pets \& Animals, Stock and Other.

\subsection{Grounding annotation}

\hspace{\parindent}\textbf{IRRELEVANT.} The snippet does not contain information relevant to the question and, therefore, cannot be utilized to generate an answer. The content of the snippet fails to address the specific query, lacking any pertinent details or insights that could contribute to formulating a comprehensive response.

\textbf{RELEVANT.} The snippet is directly relevant to the query or topic at hand. It contains detailed and substantial information that can significantly contribute to the answer creation. This means that the content of the snippet should directly address the query, offering pertinent insights, data, or explanations that enhance the model\'s ability to generate accurate and comprehensive responses.

\textbf{ANSWER-THE-QUESTION.} The information from the snippet is up to date and correctly answer the question. Make sure you check the date that the question was asked and date of the document (evidence date). When there are contradicting information on different snippets, execute a google search to decide the correct answer. 

\textbf{RELATED-INFORMATION.} The snippet does not directly answer the question on its own. However, it contains information relevant to the query, which can be utilized to expand a detailed response. The content provides details and context that can be integrated into a more comprehensive and thorough answer.

\textbf{OUTDATED.} The snippet contains information that is outdated and no longer current. This means the data or details presented are from a previous time period and do not reflect the most recent developments or findings related to the topic. As a result, the information may not be accurate or relevant in the present context, potentially leading to misunderstandings or incorrect conclusions if relied upon.

\textbf{UNKNOWN.} The source age is unknown and the age can\'t be inferred from the snippet.

\subsection{Answer qualities}

\hspace{\parindent}\textbf{Correctness.} A correct answer should be informative and provide valid information to the user that directly addresses the question being asked.

\textbf{Naturalness.} An answer should be natural, as in it is well formed, fluent, and grammatically correct according to standard English grammar/speaking rules. The response should be close to how humans respond to the question.

\textbf{Usefulness.} The answer should be useful and actionable that enables the customer to take an informed decision after reading the response.

\textbf{Objectiveness.} An answer should be objective, free from bias or personal opinion and based on verifiable facts that can be supported by evidence.

\textbf{Helpfulness.} The answer should be helpful, honest and cause no harm to the reader/human/customer.

\section{Prompts}
\subsection{LLM prompt}
\label{app:prompt_example}

\begin{tcolorbox}[title=Example Inference Prompt for Baseline Models,colback=gray!5!white,colframe=gray!75!black]
\tiny
A chat between a curious User and an artificial intelligence Bot. The Bot gives helpful, detailed, and polite answers to the User's questions.

In this session, the model has access to search results and a user's question, your job is to answer the user's question using only information from the search results.

Model Instructions:
\begin{itemize}
    \item You should provide concise answer to simple questions when the answer is directly contained in search results, but when comes to yes/no question, provide some details.
    \item In case the question requires multi-hop reasoning, you should find relevant information from search results and summarise the answer based on relevant information with logical reasoning.
    \item Be faithful to information provided in the searched passages and indicate that ``The model cannot find sufficient information to answer the question'' if the query cannot be answered using information in the search results.
    \item Place citations as inline markers (e.g., \%[1]\%, \%[2]\%, etc.) directly within the relevant parts of the response text. Do not include a separate citation section after the response.
    \item DO NOT USE INFORMATION THAT IS NOT IN SEARCH RESULTS!
\end{itemize}

Context: The current time is \{query\_time\}.

User: \{query\} Bot:

Resource: Search Results: \{search\_results\} Bot:

\end{tcolorbox}

\subsection{Facts Eligibility Prompt}
\label{app:eligibility_prompt}
\begin{tcolorbox}[title=Example Eligibility Prompt for the LLM judge,colback=gray!5!white,colframe=gray!75!black]
\tiny
Your mission is to judge the response from an AI model, the *test* response, calibrating your judgement using a *baseline* response. Please use the following rubric criteria to judge the responses:\\

<START OF RUBRICS>

Your task is to analyze the test response based on the criterion of ``Instruction Following''. Start your analysis with ``Analysis''.\\

**Instruction Following**

Please first list the instructions in the user query.

In general, an instruction is VERY important if it is specifically asked for in the prompt and deviates from the norm. Please highlight such specific keywords.

You should also derive the task type from the user query and include the task-specific implied instructions.

Sometimes, no instruction is available in the user query.

It is your job to infer if the instruction is to autocomplete the user query or is asking the 
LLM for follow-ups.

After listing the instructions, you should rank them in order of importance.

After that, INDEPENDENTLY check if the test response and the baseline response meet each of the instructions.

You should itemize, for each instruction, whether the response meets, partially meets, or does not meet the requirement, using reasoning.

You should start reasoning first before reaching a conclusion about whether the response satisfies the requirement.

Citing examples while reasoning is preferred.\\

Reflect on your answer and consider the possibility that you are wrong.

If you are wrong, explain clearly what needs to be clarified, improved, or changed in the rubric criteria and guidelines.\\

In the end, express your final verdict as one of the following three json objects:\\

\{\{
  ``Instruction Following'': ``No Issues''
\}\}\\

\{\{
  ``Instruction Following'': ``Minor Issue(s)''
\}\}\\

\{\{
  ``Instruction Following'': ``Major Issue(s)''
\}\}\\

<END OF RUBRICS>\\

\# Your task

\#\# User query

$<|begin\_of\_query|>$

\{query\}

$<|end\_of\_query|>$\\

\#\# Test Response:

$<|begin\_of\_test\_response|>$

\{answer\}

$<|end\_of\_test\_response|>$\\

\#\# Baseline Response:

$<|begin\_of\_baseline\_response|>$

\{expected\_answer\}

$<|end\_of\_baseline\_response|>$\\

Please write your analysis and final verdict for the test response.
\end{tcolorbox}

\subsection{Facts Factuality Prompt}
\label{app:grounding_prompt}
\begin{tcolorbox}[title=Example Factuality Prompt for the LLM judge,colback=gray!5!white,colframe=gray!75!black]
\tiny
You are a helpful and harmless AI assistant. You will be provided with a textual context and a model-generated response.Your task is to analyze the response sentence by sentence and classify each sentence according to its relationship with the provided context.\\

**Instructions:**\\

1. **Decompose the response into individual sentences.**

2. **For each sentence, assign one of the following labels:**

    * **`supported`**: The sentence is entailed by the given context.  Provide a supporting excerpt from the context. The supporting except must *fully* entail the sentence. If you need to cite multiple supporting excepts, simply concatenate them.

    * **`unsupported`**: The sentence is not entailed by the given context. No excerpt is needed for this label.
    
    * **`contradictory`**: The sentence is falsified by the given context. Provide a contradicting excerpt from the context.
    
    * **`no\_rad`**: The sentence does not require factual attribution (e.g., opinions, greetings, questions, disclaimers).  No excerpt is needed for this label.

3. **For each label, provide a short rationale explaining your decision.**  The rationale should be separate from the excerpt.

4. **Be very strict with your `supported` and `contradictory` decisions.** Unless you can find straightforward, indisputable evidence excerpts *in the context* that a sentence is `supported` or `contradictory`, consider it `unsupported`. You should not employ world knowledge unless it is truly trivial.\\

**Input Format:**\\

The input will consist of two parts, clearly separated:\\

* **Context:**  The textual context used to generate the response.

* **Response:** The model-generated response to be analyzed.\\

**Output Format:**\\

Your output should be in json format as follows:\\

The key should be `"grounding\_quality"` and the value should be a list of json objects with an object for each sentence in the response, containing the following fields:\\

* `"sentence"`: The sentence being analyzed.

* `"label"`: One of `supported`, `unsupported`, `contradictory`, or `no\_rad`.

* `"rationale"`: A brief explanation for the assigned label.

* `"excerpt"`:  A relevant excerpt from the context. Only required for `supported` and `contradictory` labels.\\

**Example:**\\

**Input:**\\

```
Context: Apples are red fruits. Bananas are yellow fruits.

Response: Apples are red. Bananas are green. Bananas are cheaper than apples. Enjoy your fruit!
```\\

**Output:**\\

\{\{"grounding\_quality": 

[\{\{"sentence": "Apples are red.", "label": "supported", "rationale": "The context explicitly states that apples are red.", "excerpt": "Apples are red fruits."\}\},

\{\{"sentence": "Bananas are green.", "label": "contradictory", "rationale": "The context states that bananas are yellow, not green.", "excerpt": "Bananas are yellow fruits."\}\},

\{\{"sentence": "Bananas are cheaper than apples.", "label": "unsupported", "rationale": "The context does not mention the price of bananas or apples.", "excerpt": null\}\},

\{\{"sentence": "Enjoy your fruit!", "label": "no\_rad", "rationale": "This is a general expression and does not require factual attribution.", "excerpt": null\}\}]

\}\}\\

**Now, please analyze the following context and response:**\\

**User Query:**
\{query\}\\

**Context:**
\{context\_document\}\\

**Response:**
\{answer\}
\end{tcolorbox}

\subsection{Deflection Judge Prompt}
\label{app:deflection_prompt}
\begin{tcolorbox}[title=Example Deflection Prompt for the LLM judge,colback=gray!5!white,colframe=gray!75!black]
\tiny

You are a human expert in analyzing responses to questions given by a model. \\

Your task to assign the response into exactly one of the following categories:\\

- missing: the response indicates that it is not sure, does not know the answer, or states that it cannot provide an answer.

- attempted: the response attempts to answer the question.\\

QUESTION: \{query\}\\

MODEL RESPONSE: \{answer\}\\

CATEGORY:\\

Your response should be in json format as follows:

\{\{

    "justification": (Explain why the MODEL RESPONSE is missing or attempted. Use one or two sentences maximum. Keep the answer as 
    concise as possible.)
    
    "grade": (missing or attempted),

\}\}

\end{tcolorbox}

\subsection{Search the Web}
\label{app:pipeline_questions_prompt1}
The LLM is first prompt to generate research (Prompt \textbf{Research Tasks}) tasks that are then expanded to search queries (\textbf{Prompt Creating Search Queries}).

\begin{tcolorbox}[title=Research Tasks,colback=gray!5!white,colframe=gray!75!black]
\tiny

You are an AI assistant tasked with breaking down a given question into a list of research tasks. Your goal is to help organize the research approach and ensure all necessary information is gathered to answer the question effectively.
\\ \\
Here's the question you need to analyze: \\ \\
<question> \\ 
Create complex questions that require reasoning about \{seed\_topic\}. Search the Web for recent news and documents about {{seed\_topic}} to be used as input for this creation. Those queries you will generate must require an answer that needs reasoning over evidence from multiple documents. \\
</question> \\
\\
Carefully analyze the question to identify the key components that require research. Consider the following: \\
1. The main topic or subject of the question \\
2. Any specific details, dates, or names mentioned \\
3. The type of information needed (e.g., factual, comparative, historical) \\
4. Any implicit sub-questions within the main question \\
\\
Based on your analysis, create a list of research tasks. Each task should: \\
1. Be clear, specific, and concise \\
2. Focus on one aspect of the research \\
3. Be actionable and help in gathering relevant information \\
4. Be ordered logically, if there's a natural sequence to the research \\
5. Not use pronouns to refer to something in the question. Use its actual name. \\
\\
The number of tasks can range from 1 to 3, depending on the complexity of the question. Simpler questions may only require one or two tasks, while more complex ones might need up to three. \\
\\
Present your list of research tasks in the following format: \\
<research\_tasks> \\
1. [First research task] \\
2. [Second research task] \\
3. [Third research task] \\
 ...
</research\_tasks> \\
\\
If a subsequent task specifically refers to an entity or other information that a preceding task can provide, mark that entity or item with a note in parentheses that indicates the task number that provides it. Format: (from task N) \\
Make sure to place the parenthesis inline in the text after the entity. 
Example: \\
1. Identify the name of ... \\
2. Find the ... of that person (from task 1). \\
It is important to use the correct format. If entity X was identified in task 1 and used in task 2, say X (from task 1). \\
\\

\end{tcolorbox}

\begin{tcolorbox}[title=Creating Search Queries,colback=gray!5!white,colframe=gray!75!black]
\tiny

You are an AI research assistant tasked with understanding and researching a given task. Your goal is to analyze the task and generate appropriate Web search queries for further research.
\\
Here is the task you need to research:\\
<task>\\
\{instruction\}\\
</task>\\

Follow these steps to complete your assignment:\\
\\
1. Carefully read and analyze the given task and the results of related tasks.
\\
2. Think about the key aspects of the task that require further research. Consider what information you need to gather to fully address the task.
\\
3. Based on your analysis, generate three Web search queries that will help you research the task effectively. Each query should focus on a different aspect of the task or seek different types of information.
\\
4. Today's date is {{ date }}. You may use the current year in your Web search queries where appropriate. 
\\
5. Format your output as follows:
   a. First, provide a brief explanation of your understanding of the task and how you've incorporated information from related tasks. Write this explanation inside <task\_analysis> tags.
   b. Then, list your three Web search queries, each wrapped in <QUERY> tags.
\\
Here's an example of how your output should be structured:
\\\\
<task\_analysis> [Your analysis of the task] 
</task\_analysis>\\
\\
<QUERY>[Your first Web search query]</QUERY>\\
<QUERY>[Your second Web search query]</QUERY>\\
<QUERY>[Your third Web search query]</QUERY>\\
\\
Remember, your queries should be specific and targeted to gather the most relevant information for the task at hand. Avoid overly broad or vague queries.\\

\end{tcolorbox}

\subsection{Generate Questions}
\label{app:pipeline_questions_prompt2}

\begin{tcolorbox}[title=Generating Complex Questions,colback=gray!5!white,colframe=gray!75!black]
\tiny

You are tasked generating questions, based on the results to several sub-questions. Your goal is to create detailed and focused queries that will require reasoning to be answered, addressing the main instruction while incorporating key information from each sub-question.
\\
Here is the main instruction you need to address:
<main\_instruction> \\
Create complex questions that require reasoning about \{seed\_topic\}. Search the Web for recent news and documents about {{seed\_topic}} to be used as input for this creation. Those queries you will generate must require an answer that needs reasoning over evidence from multiple documents. \\
</main\_instruction> \\ \\

Below are the sub-questions and their corresponding answers: \\
<sub\_questions\_answers> \\
\{\% for task, result in prev\_results.items() \%\} \\
<task> \\
- Task \{task\}: \{result\} \\
</task> \\
\{\% endfor \%\} \\
</sub\_questions\_answers>
\\ \\
To synthesize your response, follow these steps: \\
1. Look for common themes or connections between the sub-answers. \\
2. Based on the high level topics, generate complex questions which needs reasoning to obtain answers (e.g., "how many days did Thurgood Marshall serve as a Supreme Court justice?", "how many Oscar awards did Meryl Streep win?"). \\
3. Avoid to create simple questions that can be answered by just composing and summarizing the answers. \\
4. The questions must require chaining multiple pieces of information to compose the answer (e.g., "who acted in Ang Lee’s latest movie?") \\
5. Generate a list of elaborate questions using an unordered list. \\
6. The questions should NOT have more than 20 words, they must be succinct. Don't make the generated questions too long! \\

Example of generated elaborate questions: \\ \\
what age did ferdinand magelan discovered the philippines? \\
how many grammy awards were won by the song plan b until 62nd grammy? \\
how many feet is the place with the lowest elevation in vermont? \\
when did meta announce the release of the meta quest 4? \\
what christmas song is the most streamed, all i want for christmas is you or jingle bells? \\
who portrayed the younger character, gillian lynne in phantom of the opera: behind the mask or brian blessed in prisoner of honor? \\
was the lion king the highest-grossing film of all time when it was released? \\
how many american players ranked are in the top ten of the wta? \\
the 1984 maze game devil world has an amazing soundtrack. who composed it? \\
which movie has a higher budget, poor things or bob marley one love? \\
how many films are in the hunger games film series? \\
what was the pricing of the most recent ipo for a major social media company? \\
which feid album is the one that has "normal" on it? \\
what was the total value of all mergers and acquisitions in the healthcare sector? \\
what are the top 3 movies on hbo max? \\
how many regional confederations does fédération internationale de volleyball have? \\
how many american music awards has taylor swift won throughout her career? \\
what is the age difference between angelina jolie and billy bob thornton? \\
how much is the worst performing stock, amazon? \\
how many albums has the band radiohead released that have reached number one on the uk albums chart? \\
who was the first actress to play the role of wonder woman in a live-action movie? \\
the three countries with the highest oil production are...? \\
which artist has been performing for longer, shakira or machine gun kelly? \\
how many times has meryl streep been nominated for an academy award for best actress? \\
which female tennis players have won all four grand slam tournaments in a single calendar year? \\
which five dow jones companies have a debt-to-equity ratio of less than 0.1? \\
give at least  3 etf funds with over 1 billion total of assets \\
which movie did jennifer lawrence co-star in with bradley cooper where she played a character with a mental health condition? \\
what are the five highest-grossing movies at the box office directed by nancy meyers? \\

\end{tcolorbox}

\section{Examples}
\label{sec:examples}
Below, we present either questions from different dataset features (e.g. datapoints requiring deflection) or we present detailed examples from the dataset with the associated annotations and a response from one of the evaluated models.

\subsection{Private (Domain-specific) questions}
Examples of questions covering different private KBs, in contrast with the more general Web questions:

\begin{itemize}
     \item \textbf{DevOps}: ``How do fractional derivatives enhance the NLS model in ocean engineering?''
     \item \textbf{SEC}: ``What are the differences between the regulatory requirements for CE marking in the EU and FDA approval in the U.S. for medical devices and nutritional products?''
     \item \textbf{Enron}: ``What were the main concerns and objections raised by PG\&E regarding the DWR's power purchase revenue requirement?''
     \item \textbf{Arxiv}: ``How do performance metrics like Peak Signal-to-Noise Ratio (PSNR) and Structural Similarity Index (SSIM) evaluate the effectiveness of image denoising algorithms, and what are their limitations?''
     \item \textbf{Web}: ``Considering the focus on experience-based events in 2025, how do the world's largest religious gatherings like the Maha Kumbh Mela compare in terms of scale and significance?''
\end{itemize}

\subsection{Questions with different levels of complexity}

\begin{itemize}
    \item \textbf{Simple}: What improvements does Amazon Bedrock's new caching feature offer?
    \item \textbf{Simple w. condition}: What is the current age of Donald Trump, and how long did he serve as the President of the United States?
    \item \textbf{Comparison}: How might Blue Origin's successful New Glenn launch impact SpaceX's dominance in the private space sector?
    \item \textbf{Post-processing heavy}: What are the implications of the Gaza ceasefire negotiations for regional stability, and how is BBC News reporting on the complexities of these talks?
    \item \textbf{Multi-Hop}: With Marina Satti's absence from the Eurovision 2025 national final due to her strained relations with ERT, what is the potential effect on Greece's participation in the upcoming contest?
    \item \textbf{Set}: What strategies might be implemented to ensure the reliability and affordability of internet services in the Far North, as outlined by the CRTC?
    \item \textbf{Aggregation}: How many albums has the band Radiohead released that have reached number one on the UK Albums Chart?
\end{itemize}

\subsection{Question covering different levels of popularity}
\begin{itemize}
    \item \textbf{Head}: What impact has Biden's decision to commute the sentences of nearly 2,500 inmates serving long prison terms for nonviolent drug offenses had on the criminal justice system and public opinion?
    \item \textbf{Torso}: How do the latest Google algorithm updates reflect the company's commitment to improving search quality and providing valuable content?
    \item \textbf{Tail}: How might the new asset management joint venture between Natixis Investment Managers and Generali impact the landscape of European finance, considering the competitive environment and recent trends?
\end{itemize}

\subsection{Question with different degree of temporal dynamism}
\begin{itemize}
    \item \textbf{Evergreen}: How did Abbott Laboratories justify the acquisition cost of Visiogen, Inc. in 2009?
    \item \textbf{Slow-Changing}: How has the COVID-19 pandemic impacted Matson's operations and financial performance?
    \item \textbf{Fast-Changing}: Which two quarterbacks threw for more than 100 yards in the NFL playoffs?
\end{itemize}

\subsection{GaRAGe detailed examples}
\begin{tcolorbox}[title=GaRAGe example 1, colback=gray!5!white, colframe=gray!75!black]
\small

\subparagraph{Question:} What trends are shaping RSU compensation in 2024?\\

\begin{itemize}
    \item temporal dynamism: Slow-Changing
    \item complexity: Set
    \item domain: Stock
    \item popularity: Tail
\end{itemize}

\subparagraph{Grounding:}

\begin{itemize}
    \item Snippet 1:
        \begin{itemize}
            \item text: RSUs as a Component of a Competitive Compensation Package RSUs are often included as part of a competitive compensation package, along with base salary, bonuses, and other benefits.RSUs can enhance the overall attractiveness of a compen...
            \item age: 11 months and 8 days
            \item label: IRRELEVANT
        \end{itemize}
    \item Snippet 2:
        \begin{itemize}
            \item text: Toggle Employee stock benefit schemes—such as Restricted Stock Units (RSUs), Employee Stock Option Plans (ESOPs), and Employee Stock Purchase Plans (ESPPs) - are increasingly....
            \item age: 1 weeks and 2 days
            \item label: IRRELEVANT
        \end{itemize}
    \item Snippet 10:
        \begin{itemize}
            \item text: As Artificial Intelligence (AI) continues to transform various aspects of business, its integration with RSU compensation is becoming an emerging trend. In the 2024 fiscal year, companies are looking to leverage AI-based applications in RSU plans in the following ways:
            \item age: 1 years, 1 months, and 12 days
            \item label: RELEVANT
        \end{itemize}
\end{itemize}

\subparagraph{Human answer:} In 2024, several trends are shaping RSU compensation. Companies are increasingly integrating Artificial Intelligence (AI) into RSU plans, using AI-based applications to improve the design, distribution, and management of RSU programs [10]\\

\subparagraph{Nova Pro answer:} The integration of AI-based applications in RSU plans is an emerging trend shaping RSU compensation in 2024 [10].

\end{tcolorbox}

\begin{tcolorbox}[title=GaRAGe example 2, colback=gray!5!white, colframe=gray!75!black]
\small

\subparagraph{Question:} How do the recent findings on malaria outbreaks in the DRC affect global health strategies?\\

\begin{itemize}
    \item temporal dynamism: Slow-Changing
    \item complexity: Post-processing heavy
    \item domain: Health
    \item popularity: Torso
\end{itemize}

\subparagraph{Grounding:}

\begin{itemize}
    \item Snippet 1:
        \begin{itemize}
            \item text: In line with the group's findings, WHO updated its recommendations for 3 key malaria prevention strategies: seasonal malaria chemoprevention (SMC), ...
            \item age: 2 years, 0 months, and 13 days
            \item label: IRRELEVANT
        \end{itemize}
    \item Snippet 2:
        \begin{itemize}
            \item text: In-depth exploration is required to inform the implementation of the RTS,S vaccine particularly in countries with low or moderate EPI vaccine coverage...
            \item age: 6 months and 22 days
            \item label: IRRELEVANT
        \end{itemize}
\end{itemize}

\subparagraph{Human answer:} There is not enough grounding for an answer.\\

\subparagraph{Qwen 32b answer:}The recent findings on malaria outbreaks in the Democratic Republic of the Congo (DRC) have influenced global health strategies by prompting the World Health Organization (WHO) to update its recommendations for malaria prevention strategies. These strategies include seasonal malaria chemoprevention (SMC), perennial malaria chemoprevention (PMC), and intermittent preventive treatment in pregnancy (IPTp). [1] The updates aim to broaden the use of chemoprevention among young children at high risk and to expand access to preventive chemotherapy during pregnancy. These recommendations reflect a more informed approach to when and where chemoprevention is most effective, encouraging tailored deployment in local contexts. [3] Additionally, the WHO is developing a Framework to guide decisions on the allocation of limited malaria vaccine supplies, ensuring that national malaria control programmes can make informed decisions about vaccine prioritization. [4] These adjustments in global health strategies are driven by the latest evidence and aim to enhance the effectiveness of malaria control measures, particularly in high-risk areas like the DRC. [1], [2], [3], [4]

\end{tcolorbox}

\end{document}